\documentclass{article}

\PassOptionsToPackage{numbers, compress}{natbib}
\usepackage[dvipsnames]{xcolor}

\definecolor{myblue}{RGB}{31, 119, 180}
\definecolor{myorange}{RGB}{255, 127, 14}
\definecolor{mygreen}{RGB}{44, 160, 44}
\definecolor{myred}{RGB}{214, 39, 40}
\definecolor{myyellow}{RGB}{230,194,0}

\usepackage[final]{neurips_ai4d3_2025}

\usepackage[utf8]{inputenc} %
\usepackage[T1]{fontenc}    %
\usepackage[allcolors=NavyBlue,colorlinks=true,backref=page]{hyperref}       %
\usepackage{url}            %
\usepackage{booktabs}       %
\usepackage{amsfonts}       %
\usepackage{nicefrac}       %
\usepackage{microtype}      %
\usepackage{xspace}
\usepackage{subcaption}

\usepackage{wrapfig}

\usepackage[noend]{algpseudocode}
\usepackage{algorithm}

\usepackage{cite}
\usepackage{comment}
\usepackage{prettyref}
\usepackage{amsfonts}
\usepackage{makecell}
\usepackage{adjustbox}
\usepackage{multicol}

\usepackage{amsmath}
\usepackage{amssymb}
\usepackage{mathtools}
\usepackage{amsthm}

\usepackage[textsize=footnotesize]{todonotes}
\usepackage{xspace}

\definecolor{colorcomment}{RGB}{160, 190, 210}%
\makeatletter
\algnewcommand{\LineComment}[1]{\Statex \hskip\ALG@thistlm \(\triangleright\) 
{\color{colorcomment}#1}}
\makeatother

\makeatletter
\algnewcommand{\IndentLineComment}[1]{\Statex \hskip\ALG@tlm \(\triangleright\) {\color{colorcomment}#1}}
\makeatother

\newcommand{\given}{{\,|\,}}
\newcommand{\Prob}{{P}}

\newcommand\policy{\ensuremath{\pi}}

\newcommand\stateDist{d}

\newcommand\RFunc{R}

\newcommand\horizon{\ensuremath{T}}
\newcommand\ENum{\ensuremath{N}} %

\newcommand\dataset{\ensuremath{\mathcal{D}}}

\newcommand\critic{\ensuremath{{C}}\xspace}

\newcommand{\expctover}[2]{\mathbb{E}_{#1}\!\left[#2\right]}
\newcommand\RationaleBuffer{\ensuremath{\mathcal{B}}}

\def \argmax {\mathop{\rm arg\,max}}

\newcommand\Gen{\ensuremath{\policy^{{G}}_{{\theta}}}}

\newcommand\Vocabulary{\ensuremath{\mathcal{V}}}

\newcommand{\hypothesis}{\mathcal{Y}}

\newcommand\Buffer{\ensuremath{\mathcal{B}}}

\newcommand\corpus{\ensuremath{C}}

\newcommand\tokenSeq{\ensuremath{{F}}}
\newcommand\token{\ensuremath{{f}}}
\newcommand{\algname}{\textsc{FragmentGPT}\xspace}

\newif\iffinal
\finaltrue

\iffinal
    \newcommand{\fix}[1]{#1}
    \newcommand{\note}[1]{}
    \newcommand{\pref}[1]{}
    \newcommand{\XL}[1]{}
    \newcommand{\SJ}[1]{}
    \newcommand{\XLinline}[1]{}
\else
    \newcommand{\fix}[1]{{\color{red} #1}}
    \newcommand{\SJ}[1]{\todo[fancyline,color=Blue!40]{SJ: #1}\xspace}
    
    \newcommand{\XL}[1]{\todo[fancyline,color=Maroon!40]{XL: #1}\xspace}
    \newcommand{\XLinline}[1]{\textcolor{Maroon}{[XL: #1]}}
    
    \newcommand{\note}[1]{{\color{purple}[XL: #1]}}
    
    \newcommand{\pref}[1]{{\color{blue}(\ref{#1})}}
\fi

\newcommand{\tabref}[1]{Table~\ref{#1}}

\newcommand{\secref}[1]{\S\ref{#1}}
\newcommand{\appref}[1]{Appendix~\ref{#1}}

\newcommand{\paren} [1] {\ensuremath{ \left( {#1} \right) }}

\newcommand{\bracket}[1]{\left[#1\right]}
\newcommand{\tuple}[1]{\ensuremath{\left\langle #1 \right\rangle}}
\newcommand{\curlybracket}[1]{\ensuremath{\left\{#1\right\}}}

\theoremstyle{plain}

\theoremstyle{definition}

\theoremstyle{remark}

\title{
\algname: A Unified GPT Model for Fragment Growing, Linking, and Merging in Molecular Design
}

\author{Xuefeng Liu\textsuperscript{1,4}\thanks{Equal Contribution. Correspondence to: Xuefeng Liu <\href{mailto:xuefeng@uchicago.edu}{xuefeng@uchicago.edu}>.} ,~\textbf{Songhao Jiang\textsuperscript{1$\ast$}},~\textbf{Qinan Huang\textsuperscript{2}},~\textbf{Tinson Xu\textsuperscript{1}}\\ \textbf{Ian Foster\textsuperscript{1,4}},~\textbf{Mengdi Wang\textsuperscript{5}},~\textbf{Hening Lin\textsuperscript{3}},~\textbf{Rick Stevens\textsuperscript{1,4}} \\
\textsuperscript{1}Department of Computer Science, University of Chicago\\
\textsuperscript{2}Pritzker School of Molecular Engineering, University of Chicago\\
\textsuperscript{3}Department of Chemistry, University of Chicago\\
\textsuperscript{4}Argonne National Laboratory\\ 
\textsuperscript{5}AI Lab, Princeton University\\
}

\begin{document}

\maketitle

\begin{abstract}
Fragment-Based Drug Discovery (FBDD) is a popular approach in early drug development, but designing effective linkers to combine disconnected molecular fragments into chemically and pharmacologically viable candidates remains challenging. Further complexity arises when fragments contain structural redundancies, like duplicate rings, which cannot be addressed by simply adding or removing atoms or bonds. 
To address these challenges in a unified framework, we introduce \algname, which integrates two core components: (1) a novel chemically-aware, energy-based bond cleavage pre-training strategy that equips the GPT-based model with fragment growing, linking, and merging capabilities, and (2) a novel Reward Ranked Alignment with Expert Exploration (RAE) algorithm that combines expert imitation learning for diversity enhancement, data selection and augmentation for Pareto and composite score optimality, and Supervised Fine-Tuning (SFT) to align the learner policy with multi-objective goals.
Conditioned on fragment pairs, \algname generates linkers that connect diverse molecular subunits while simultaneously optimizing for multiple pharmaceutical goals. It also learns to resolve structural redundancies—such as duplicated fragments—through intelligent merging, enabling the synthesis of optimized molecules. \algname facilitates controlled, goal-driven molecular assembly.  Experiments and ablation studies on real-world cancer datasets demonstrate its ability to generate chemically valid, high-quality molecules tailored for downstream drug discovery tasks.

\end{abstract} 

\section{Introduction}

Fragment-based drug design (FBDD) has become a mainstream strategy in early drug discovery, especially for challenging targets like protein–protein interactions, allosteric sites, and shallow or cryptic pockets~\citep{weijunxu2025}. Unlike high-throughput screening (HTS), which depends on large, complex compound libraries~\citep{LEACH2011489}, FBDD employs small, low-molecular-weight fragments to probe binding surfaces with precision and efficiency~\citep{Keserű2016,LEACH2011489,weijunxu2025,murray2009rise}. Though fragments typically bind weakly (micromolar–millimolar range), they offer high ligand efficiency and chemical tractability, making them ideal starting points for optimization~\citep{wasko2015fragment}. In addition, fragment libraries, usually only hundreds to a few thousand compounds, cover chemical space more effectively than HTS due to the combinatorial potential of linking or growing fragments. Their small, hydrophilic nature often yields hits with favorable solubility and reduced off-target risks. Moreover, sensitive biophysical methods such as NMR, X-ray crystallography, SPR, and thermal shift assays enable rapid screening and validation, making FBDD a powerful, efficient route for lead discovery.

The impact of FBDD is underscored by the approval of several clinically significant drugs. For example, Zelboraf (vemurafenib), a BRAF inhibitor for melanoma, was the first FDA-approved drug derived from FBDD \citep{bollag2012vemurafenib}. It was followed by others such as Venetoclax, a Bcl-2 inhibitor disrupting protein–protein interactions \citep{fairbrother2019venetoclax}, and Sotorasib, a KRAS-G12C inhibitor targeting a previously “undruggable” oncogene \citep{Lanman2020Sotorasib}. In total, eight FDA-approved drugs and over 50 clinical candidates have originated from FBDD pipelines, validating its versatility across diverse target classes, including kinases, protein–protein interactions, and even RNA-binding proteins \citep{weijunxu2025}.

Despite its success, FBDD faces notable challenges. Once a fragment hit is identified, a major challenge lies in transforming these minimal binders into high-affinity ligands while maintaining or enhancing their binding specificity and desirable physicochemical traits. This is generally accomplished through three primary optimization strategies: fragment growing, fragment merging, and fragment linking. 
 
Fragment growing involves expanding a single fragment into nearby binding regions, guided by structural information or structure–activity relationships (SAR). 
The clinical success of fragment growing is exemplified by several notable drug discoveries. Smith and colleagues advanced MRTX1719, targeting the PRMT5·MTA complex, beginning with 4-(aminomethyl)quinazoline-1(2H) identified through surface plasmon resonance \citep{smith2022fragment}. 
Similarly, Heinrich and colleagues developed focal adhesion kinase inhibitors starting from 1H-pyrrolo[2,3-b]pyridine 
scaffolds \citep{heinrich2013fragment}, 
while Saxty et al. utilized their "Pyramid" X-ray crystallography platform to evolve 5-methyl-4-phenyl-1H-pyrazole hits into potent protein kinase B inhibitors \citep{wyatt2008identification}. Likewise, Addie and colleagues at AstraZeneca started with a small hinge-binding pyrrolopyrimidine fragment and, using crystal structures to guide each tweak, built it into AZD5363 (capivasertib)—an oral, ATP-competitive Akt inhibitor with broad isoform potency, better ROCK selectivity, low hERG liability, and strong anti-tumor activity in cells and animals \citep{addie2013astra}. 
Fragment growing offers distinct advantages including efficient exploration of chemical space through incremental expansion, structure-guided optimization based on detailed binding insights, and typically higher ligand efficiency compared to traditional high-throughput screening approaches. The method enables rational design decisions by leveraging crystallographic or NMR structural information to direct growth toward productive binding site regions.

However, fragment growing presents significant challenges that must be carefully addressed. Primary considerations include determining optimal growth directions to maximize binding affinity while avoiding steric conflicts, maintaining geometric compatibility between added fragments and binding sites, and preserving drug-like properties throughout the expansion process. Additionally, ensuring synthetic accessibility of designed compounds remains crucial for practical development timelines and costs.
The success of fragment growing depends on balancing multiple competing objectives: enhancing binding affinity and selectivity while maintaining favorable pharmacokinetic properties, drug-likeness, and synthetic feasibility. This multi-parameter optimization requires integration of structural biology, medicinal chemistry expertise, and sophisticated computational design tools to navigate the complex structure-activity relationships that emerge as fragments evolve into lead compounds.

Fragment merging integrates two or more overlapping fragments with shared substructures or binding features to preserve favorable interactions and enhance potency. Traditionally, this requires crystal structures to pinpoint overlaps—a labor-intensive process demanding expert knowledge—and even then, maintaining bioactive conformations often necessitates extensive optimization. Successful examples include trypanothione reductase inhibitors, where a propylphenyl and p-fluorophenyl group were merged via a piperazine scaffold~\citep{Exertier2024}, and HSP90 inhibitors, where adjacent fragments sharing a phenol moiety were combined but initially showed weak activity due to desolvation penalties~\citep{Ren2014HSP90}. Computational tools like the Fragment Network extend this idea by searching over 120 million catalogue compounds to suggest merges beyond simple fingerprint similarity~\citep{Wills2023}, offering access to new regions of chemical space. However, such methods remain constrained by database coverage and lack true generative flexibility, leaving fragment merging an appealing yet underdeveloped strategy in practice.

Fragment linking connects two distinct fragments, each binding in adjacent pockets, using a chemical linker. Although both merging and linking offer conceptual advantages, they are notoriously challenging to implement: merging demands precise alignment of shared pharmacophores and often relies on high-resolution co-crystal structures, while linking must address geometric limitations and design appropriate linkers to avoid entropic losses and ensure both fragments contribute to binding. These strategies frequently necessitates iterative synthesis guided by expert intuition. The increasing size of fragment libraries and growing demand for rapid optimization further highlight the need for computational tools—including virtual screening and AI-driven design—to enhance hit identification and linker optimization.

Recent advances in machine learning have introduced generative models as powerful tools for automated linker design. Sequence-based models, such as SyntaLinker \citep{yang2020syntalinker}, formulate linker generation as a sequence-to-sequence task, treating linkers as SMILES strings and employing natural language processing techniques, particularly Transformer-based neural networks. These models offer fast inference and strong chemical priors, effectively capturing the syntactic and semantic regularities of chemical representations. However, their reliance on 1D chemical language often necessitates post-processing steps to ensure the 3D structural feasibility and compatibility of generated linkers with protein binding pockets. In contrast, 3D geometry-based models, including those based on diffusion processes and variational autoencoders (VAEs)—such as 3DLinker \citep{huang20223dlinker}, LinkerNet \citep{guan2023linkernet}, DiffLinker \citep{igashov2024difflinker}, and DiffPROTACs \citep{li2024diffprotacs} —generate linkers directly in three-dimensional space. These models exploit E(3)-equivariant architectures that inherently preserve rotational and translational symmetries, enabling the generation of spatially compatible and structurally plausible linkers conditioned on the geometric context of fragment binding poses and protein pockets.

In particular, DiffLinker, a state-of-the-art E(3)-equivariant 3D-conditional diffusion model, has demonstrated impressive performance in generating chemically valid, diverse, and pocket-compatible linkers between arbitrary numbers of fragments \citep{igashov2024difflinker}. By modeling atomic geometries directly and conditioning on both fragment coordinates and protein pockets, DiffLinker surpasses prior approaches based on SMILES or autoregressive graph construction in terms of synthetic accessibility, drug-likeness, and  similarity of samples to reference compounds. However, despite these achievements, current methods—including DiffLinker—focus solely on linking distinct fragments and remain limited in handling fragment merging, where two overlapping or partially redundant fragments must be merged into a single coherent substructure. No existing generative framework can yet simultaneously address both the linking and merging tasks in fragment-based design. Developing such unified strategies is critical for enabling more flexible and realistic drug design workflows where overlapping binding poses or scaffold overlaps frequently occur.

Currently, no generative framework can jointly address both fragment linking and merging and growing. The rise of large language models (LLMs) introduces a powerful paradigm for generative chemistry, yet their potential in fragment-based design—particularly linking and merging—remains largely underexplored. 
Moreover, pretrained GPT models rely solely on maximum likelihood training, lacking explicit objective-driven optimization. 
What's more, pretrained GPT models have shown encouraging results in molecular tasks, highlighting the opportunity to harness these expert models~\citep{liu2022cost,liu2023blending} to boost the performance of our framework.
Finally, biomolecular design is inherently a multi-objective optimization problem. Multi-property evaluation measures how well a model balances multiple, often conflicting, objectives at once. One approach is a composite score, where different property metrics (e.g., LogP, QED, synthetic accessibility) are combined into a single weighted objective, guiding optimization methods. Another is Pareto optimality, where solutions are considered optimal if no property can be improved without worsening another; the collection of these solutions forms the Pareto front, which helps visualize trade-offs among objectives.
However, both approaches come with their own advantages and limitations. To address these challenges, we asked

\begin{center}
\textit{How can language models be employed to achieve fragment growing, linking, and merging, while integrating chemically-aware and expert knowledge to improve diversity under Pareto and composite score optimality alignment?
}
\end{center}

In this work, we introduce the first GPT-based framework that unifies fragment growing, linking, and merging within a single language model. Our method incorporates a novel pretraining strategy infused with chemically-aware and energy-based knowledge, allowing the model to execute these tasks without relying on pre-existing fragment databases. Instead, it directly identifies linkers and mergeable substructures from the molecular context.
To further guide generation, we propose a Reward Ranked Alignment with Expert Exploration (RAE) algorithm that combines expert imitation learning with supervised fine-tuning. Expert imitation learning, leveraging pretrained expert models, promotes diversity, while supervised fine-tuning—together with data selection and augmentation—drives the learner policy toward Pareto-optimal and composite score solutions in multi-objective molecular property optimization.
Taken together, the integration of GPT-driven fragment pretraining, expert-guided imitation, and targeted data augmentation establishes a scalable, adaptable, and goal-oriented framework that surpasses conventional database-driven approaches.

\section{Preliminaries}

\textbf{LM.} Let $\tokenSeq=\bracket{\token_1,\token_2,\cdots,\token_n}$ denote an input sequence of tokens (fragment-based prompt), where each $\token_i$ belongs to a vocabulary $\Vocabulary$. The output sequence is defined as $M=\bracket{m_1,m_2,\cdots,m_T}$ with $m_i\in\mathcal{M}$, where $\Vocabulary$ and $\mathcal{M}$ may be distinct vocabularies. For convenience, we write $\mathbf{m}{<t}=\bracket{m_1,\cdots,m{t-1}}$ and $\mathbf{m}_{T}:=M$, and use $\horizon$ to denote sequence length.
Each training sample begins with a start token $\bracket{\text{BOS}}$, followed by a sequence $\mathbf{m}$ with $y_i\in\Vocabulary$, and terminates with an end token $\bracket{\text{EOS}}$. Molecules are represented as token sequences $\mathbf{m}$ that assemble into SMILES strings, covering both partial and complete structures. We denote string concatenation by $\circ$ and the Kleene closure of $\mathcal{V}$ by $\mathcal{V}^*$. The training corpus $\corpus$ is thus:
    $\corpus := \curlybracket{\text{[BOS]} \circ \mathbf{v} \circ \text{[EOS]}~|~\mathbf{v}\in \mathcal{V}^*}.$
The generator policy $\policy_{\theta}$, parameterized by a deep neural network with weights $\theta$, defines a distribution over sequences as
$\policy_{\theta}\paren{\mathbf{m}|\mathbf{f}}=\prod_{t=1}^{|\mathbf{m}|} \policy_{\theta}\paren{m_t|\mathbf{f},\mathbf{m}_{<t}}$, where $\policy_{\theta}(m_t|\mathbf{f},\mathbf{m}{<t}) = P(m_t|\mathbf{m}{<t},X)$ denotes the conditional probability of token $m_t$.
Decoding aims to select the most likely sequence from the hypothesis space: $\mathbf{m}^{\star}=\argmax_{\mathbf{m}\in \hypothesis_{\horizon}} \log \policy_{\theta}\paren{\mathbf{m}|\mathbf{f}}. $ Causal language modeling (CLM) reformulates the training objective as $\max_{\theta}\sum^n_{i=1} \log P\paren{f_i|{\tokenSeq}_{<i};\theta}$, where $P\paren{f_i|{\tokenSeq}_{<i};\theta}$, where $P(f_i|\tokenSeq_{<i};\theta)$ is the conditional probability of token $\token_i$ given its history. In this work, we employ a GPT policy $\policy_{\theta}$ trained on dataset $\Buffer$ to capture prior chemical knowledge and generate chemically valid molecules. Molecules are represented using SMILES strings \citep{weininger1988smiles}. The tokenizer is first pre-trained with the Byte Pair Encoding (BPE) algorithm \citep{gage1994new_bpe, sennrich2015neural_bpe}. Building on this, we adopt an incremental pre-training strategy inspired by ScaffoldGPT \citep{liu2025scaffoldgpt}, which significantly enhances the model’s ability to reconstruct valid molecules from fragments.

\textbf{Fragment-based Drug Discovery.} Given an initial prompt (fragment candidates) $\curlybracket{\tokenSeq_{i=1}^n}\sim \mathcal{X}$ and a drug discovery policy $\policy_{\theta}$, the goal in fragment-based drug discovery is to find the optimal policy $\policy_{\theta^*}$ that maximizes the following objective:
\begin{equation}\label{eq:drug_generation_objective}
\policy_{\theta^*}=
\argmax_{\policy_{\theta}}
\expctover{ \curlybracket{\tokenSeq} \sim {\stateDist_0}}{\RFunc(M)|\theta,\curlybracket{\tokenSeq_{i=1}^n}}, 
\end{equation}
where 
 $M=\policy_{\theta}\paren{\cdot\given \curlybracket{\tokenSeq_{i=1}^n}}, M_{1:\horizon}=\paren{m_1,\ldots,m_t,\ldots,m_{\horizon}},m_t\in \mathcal{V}.$
 Regards to $\curlybracket{\tokenSeq_{i=1}^n}$,
For $\curlybracket{\tokenSeq_{i=1}^n}$, ${F}_1$ corresponds to fragment growing, ${F}_1^2$ corresponds to fragment linking and merging, and ${F}_1^n$ with $n>2$ corresponds to the linking and merging of multiple fragments.

\begin{figure*}[t]
        \centering
        \includegraphics[%
        width=14cm, 
        clip={0,0,0,0}]{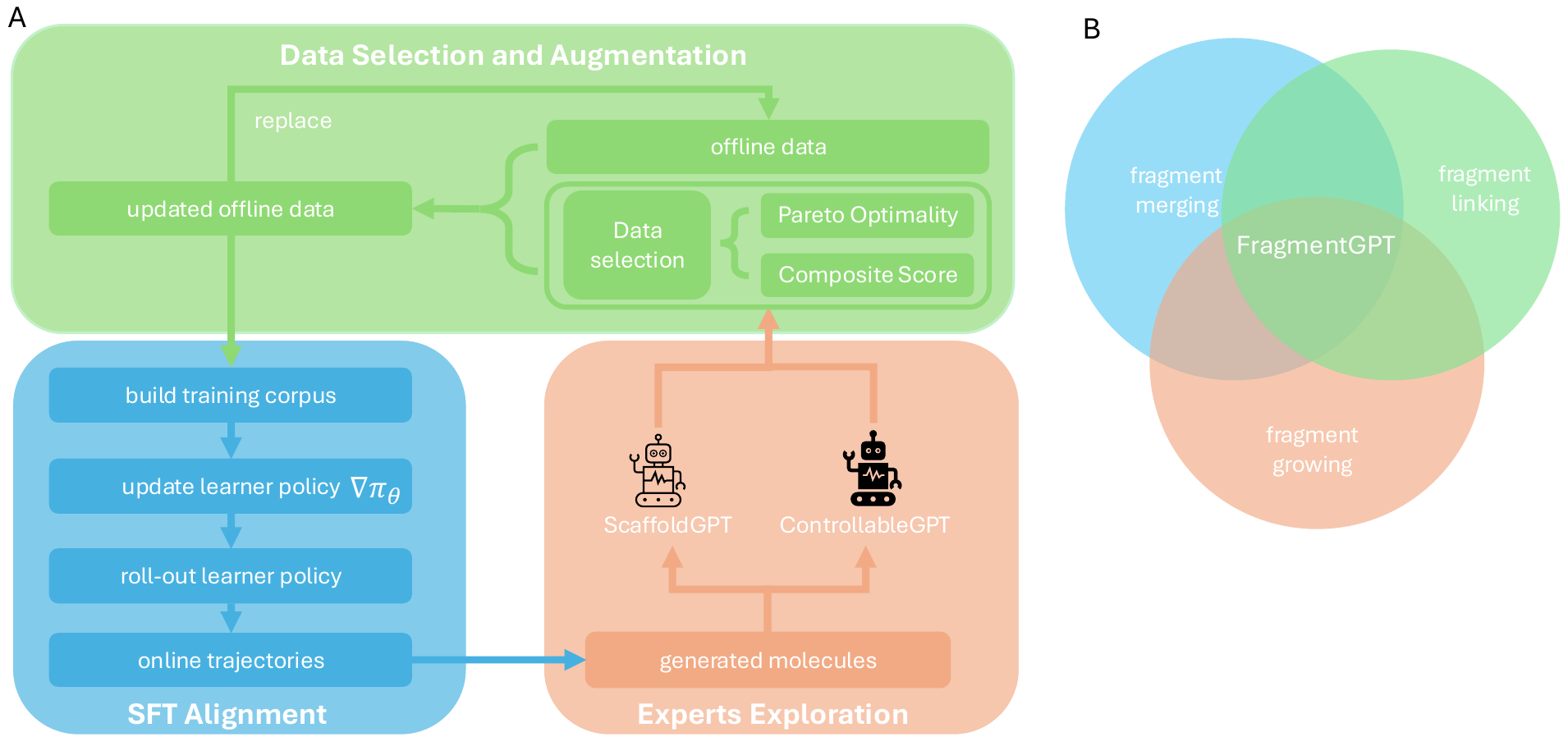}
    \caption{(A) The framework of Reward Ranked Alignment with Expert Exploration (RAE). (B) Functional overview of \algname, which supports fragment merging, linking, and growing.
    }
  \label{fig:scaffoldgpt_overview}
\end{figure*}

\section{The \algname Algorithm}\label{alg}

In this work, we address this gap by directly training a GPT model on SMILES strings, enabling it to generate complete molecules from fragment-based prompts. To achieve this, we introduce RAE (Reward Ranked Alignment with Expert Exploration), a framework that integrates SFT retraining, expert-guided exploration, and data selection and augmentation.

\begin{figure}
    \centering
    \includegraphics[width=1\linewidth]{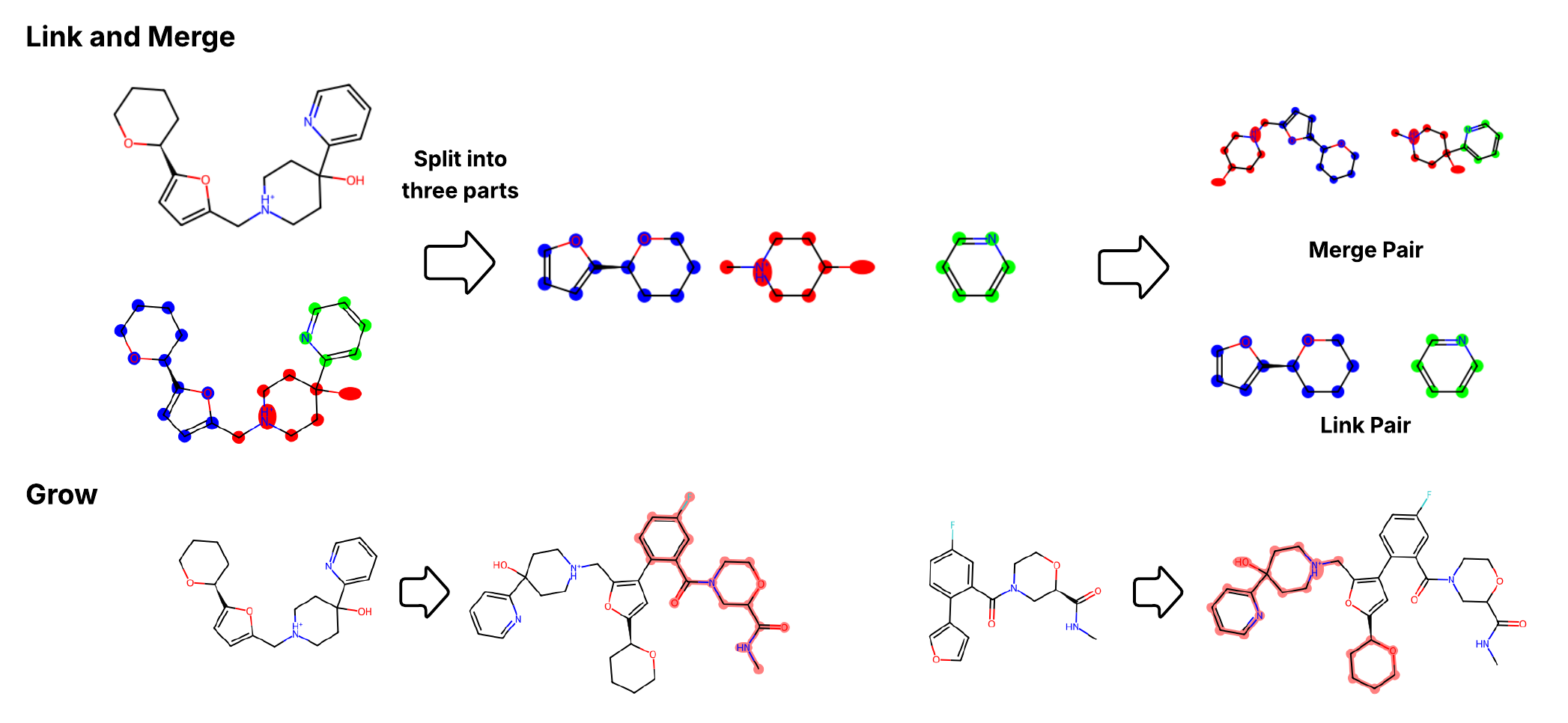}
    \caption{Overview of training corpus construction for linking, merging and growing. For linking and merging, given a SMILES string, we divide it into three parts, which we denote as fragments $\color{blue}A$, $\color{red}B$, 
 and $\color{green}C$. For the construction of the \textit{linker training corpus}, we use fragments $A$ and $C$ as inputs, and represent each training instance as <p1>$\textsc{SMILES}_{\color{blue}A}$<p2>$\textsc{SMILES}_{\color{green}C}$<L>$\textsc{SMILES}_{\color{blue}A\color{red}B\color{green}C}$. Similarly, for the \textit{merging training corpus}, we take the concatenated fragments $AB$ and $BC$ as inputs, resulting in the representation  <p1>$\textsc{SMILES}_{\color{blue}A\color{red}B}$<p2>$\textsc{SMILES}_{\color{red}B\color{green}C}$<L>$\textsc{SMILES}_{ABC}$. For growing given fragment $A$ and a molecule $M$ which contains $A$ as substructure, the model will take <p1>$\textsc{SMILES}_{A}$<L>$\textsc{SMILES}_{M}$}
    \label{fig:v2_corpus}
\end{figure}

\textbf{Stage 1. Reward Ranked Alignment.}

\textbf{\emph{Pre-Training Corpus Construction.}}
To enable the model to perform three key operations—linking, merging, and growing—we explicitly construct pretraining corpus that reflects these tasks.
Our approach begins by converting a database of molecules from their SMILES strings into graph representations, where atoms serve as nodes and bonds as edges. The core of our methodology is a chemically-aware, energy-based bond cleavage strategy. To ensure the generation of stable fragments, we annotate each bond with its dissociation energy and explicitly prevent the cleavage of bonds within aromatic rings. The algorithm preferentially targets weaker bonds for disconnection, guided by two key parameters: a maximum of three fragments per molecule to ensure substructures are significant, and a bond energy cutoff of 90 kcal/mol (~377 kJ/mol) to selectively cleave common single bonds while preserving stronger ones.
By applying this graph-based fragmentation algorithm across a large-scale molecular database, we have successfully generated a comprehensive corpus. This dataset is consist of logically paired and chemically stable fragments.

\emph{Linking:} After we constructed a large corpus of chemically meaningful fragments for linker and merger tasks, we generate a fragment-pair dataset.
As illustrated in Figure \ref{fig:v2_corpus}, given a molecule $M$, we partition it into three components: \textcolor{myblue}{A}, \textcolor{myred}{B}, \textcolor{mygreen}{C}. This yields two classes of fragment pairs. First, the pair $\paren{\textcolor{myblue}{A}, \textcolor{mygreen}{C}}$ where the task is to generate a linker bridging the two fragments.  Each resulting fragment pair, together with the original molecule $M$, forms a training instance for linking or merging.

\emph{Merging:} Second, the pairs $\paren{\textcolor{myblue}{A}\textcolor{myred}{B}, \textcolor{myred}{B}\textcolor{mygreen}{C}}$ where the task is to merge fragments based on their shared substructure $\textcolor{myred}{B}$. To further enrich the corpus and encourage the model to learn molecule merging, we construct additional examples by randomly selecting two molecules and merging them via their maximum common substructure (MCS). As shown in Figure \ref{fig:v2_corpus}, given two molecules $\textsc{MOL}_A$ and $\textsc{MOL}_B$, we first identify their MCS (highlighted in red), then align and merge them to form a new molecule $\textsc{MOL}_M$ The corresponding training corpus is represented as <p1>$\textsc{SMILES}_A$<p2>$\textsc{SMILES}_B$<L>$\textsc{SMILES}_M$ where the model learns to merge from two random molecules into a larger molecule.

\emph{Growing:} Growing is used to further refine a fragment or a molecule obtained from linking or merging. The pretraining corpus is constructed as 
<p1>\textsc{SMILES}$_{A \text{ or } B}$<L>\textsc{SMILES}$_{M}$, 
where \textsc{SMILES} may represent either a fragment or a complete molecule.
The objective of growing is to transform the fragment or molecule into a more complete and refined structure.

The corpus $\corpus$ and prompts $\mathcal{X}$ are constructed as follows:
\begin{align}\label{eq:phase2_corpus}
\curlybracket{
[BOS],\overbrace{\tuple{p_1},\underbrace{{f_1}_1,\cdots,{f_1}_{\horizon}}_{\text{fragment $F_1$}},\tuple{p_2},\underbrace{{f_2}_1,\cdots,{f_2}_{\horizon}}_{\text{fragment $F_2$}},\tuple{L}}^{\textsc{prompt}},\underbrace{m_1,\cdots,m_{\horizon}}_{\text{target molecule M}},[EOS]}
.
\end{align}

Consequently, the model is able to generate the corresponding complete molecule when provided with a pair of fragments as the input prompt.

\textbf{\emph{Pretrain a GPT generator.}}
We denote the GPT generator policy by $\policy_{\theta}$, which models the probability $p$ of producing the $t^{\text{th}}$ token in a target molecule $Y$. This probability is conditioned on all previously generated tokens $\mathbf{y}_{<t} = \bracket{y_1, \ldots, y_{t-1}}$ in the sequence, as well as on the given pair of fragment compounds $\curlybracket{F_{i=1}^n}$, formally expressed as
$\policy_{\theta}\paren{y_t \given \mathbf{y}_{<t},\curlybracket{F_{i=1}^n}}=p\paren{y_{t}|\mathbf{y}_{<t},\curlybracket{F_{i=1}^n}}.$
The parameters $\theta$ of the generator policy $\policy_{\theta}$ are optimized by minimizing the negative log-likelihood (NLL) over the training corpus, computed on the complete SMILES sequences across the entire dataset.
This process is described as follows:
\begin{align}\label{eq:pretrain-loss}
    NLL =&  - \log \Prob\paren{Y|\curlybracket{F_{i=1}^n}} \\
    =&-\sum_{t=1}^{\horizon} \log \Prob\paren{y_{t}|y_{t-1},...,y_{1},\curlybracket{F_{i=1}^n}}
   =-\sum_{t=1}^{\horizon} \log \policy_{\theta}\paren{y_t|y_{1:t-1}, \curlybracket{F_{i=1}^n}},
\end{align}
Here, $\horizon$ denotes the total number of tokens in the target molecule $Y$. The negative log-likelihood (NLL) thus measures the likelihood of reconstructing a designated target molecule from the given fragment pair $\curlybracket{F_{i=1}^n}$.
In this project, we adopt an incremental training strategy~\citep{liu2025scaffoldgpt} that leverages large collections of unlabeled text to build a foundational model of language understanding. This foundation model can then be fine-tuned and adapted to a variety of downstream tasks with specific objectives.

\begin{algorithm}[t]
    \caption{Reward Ranked Alignment with Expert Exploration (RAE)}\label{alg:scaffoldgpt}
    \begin{algorithmic}[1] 
    \Require { GPT-based generator policy $\Gen$; offline dataset \dataset; replay buffer \RationaleBuffer; number of experts E.}
    \State $\dataset \rightarrow \RationaleBuffer$. Initialize $\Gen$ with GPT2-like Transformer with random weight $\theta$.

    \For{$n=1,\ldots,\ENum$}

    \LineComment{/* Stage 1: SFT retraining */}
    \State $\curlybracket{M}_{i=1}^N \sim \rho_0$, { \text{where }$\rho_0\in \Delta\paren{\dataset}$}. $\RationaleBuffer \leftarrow \emptyset$.
    \State Build the prompt $\mathcal{X}$ and training corpus  $\corpus_\text{growing}$, $\corpus_\text{linking}$, $\corpus_\text{merging}$~(\ref{eq:phase2_corpus}) by using $\curlybracket{M}_{i=1}^N$. 
    \State Pre-train BPE tokenizer and $\policy_{\theta}$ on  $\corpus_\text{growing}$, $\corpus_\text{linking}$ and $\corpus_\text{merging}$ via CLM objective~\eqref{eq:pretrain-loss}.
    \State Generate $M_{i=1}^N=\curlybracket{\paren{m_t,\ldots,m_{\horizon}}}_{i=1}^N\sim \policy_{\theta}\paren{\cdot| \curlybracket{F_{i=1}^n}  }$, $\curlybracket{F_{i=1}^n}\sim \mathcal{X}$.
    \State $\RationaleBuffer\leftarrow\RationaleBuffer\cup M_{i=1}^N$.
    \LineComment{/* Stage 2: Expert Exploration */}
    \State $ \curlybracket{M^{\textsc{expert}}}_{i=1}^N \sim \policy^{\textsc{expert}}_j\paren{M_i}, i\in \bracket{N}, j\in \bracket{E}$.
    \State $\RationaleBuffer\leftarrow\RationaleBuffer\cup \curlybracket{M^{\textsc{expert}}}_{i=1}^N$.
    \LineComment{/* Stage 3: Data Selection and Augmentation */}
    \State $\dataset \leftarrow 80\% \cdot \textsc{Priority}\paren{\RationaleBuffer,\RFunc^{\textsc{Pareto}},\RFunc^{\textsc{Composite}}}~\cup~ 20\% \cdot \textsc{Random}\paren{\dataset}$.

    \EndFor
    
    \end{algorithmic}
\end{algorithm}

\textbf{Stage 2. Expert Exploration.}
We begin by sampling a batch of prompts 
$\mathcal{X}_t=\curlybracket{{\curlybracket{F_{i=1}^{n}}}_1^t,\cdots,{\curlybracket{F_{i=1}^{n}}}_b^t} \subseteq \mathcal{X}$, and for each ${\curlybracket{F_{i=1}^{n}}}_i^t\in \mathcal{X}_t$, we generate molecules $M_1,\cdots,M_{K}\sim \policy_{\theta}\paren{\cdot|{\curlybracket{F_{i=1}^{n}}}_i^t}$. Next, we leverage the expert policy $\policy^{\textsc{E}}$ to produce more diverse molecular candidates: 
$M_1^{\textsc{Expert}},\cdots,M_{K}^{\textsc{Expert}}\sim \policy^{\textsc{Expert}}\paren{\cdot|M_i}, i \in \bracket{K}$. 
Specifically, we employ ScaffoldGPT~\citep{liu2025scaffold, liu2025scaffoldgpt}  and ControllableGPT~\citep{liu2025controllablegpt,liu2025ground} to promote structural diversity. 
Finally, we combine the molecules generated by both the learner policy and the expert policy into the replay buffer $\RationaleBuffer$, $\RationaleBuffer \leftarrow \curlybracket{M} \cup \curlybracket{M^{\textsc{Expert}}}$.

\textbf{Stage 3. Data Selection and Augmentation.}
In this step, we employ the Pareto optimality reward model ${\RFunc^{\textsc{Pareto}}\paren{M}}$ from~\citet{abeer2024multi} and the composite score reward model ${\RFunc^{\textsc{Composite}}\paren{M}}$ from~\citet{liu2025scaffoldgpt}. 
In this work, we regarded each pharmaceutical property as a critic $C$ and considered an ensemble critics $\mathbf{C}$ as follows:
\begin{align*}
[\critic^{\text{Druglikeness}},
\critic^{\text{Solubility}},
\critic^{\text{Synthesizability}}, 
{\critic^{\text{Docking}}},\critic^{\text{{Tanimoto}}}
],
\end{align*}
where each critic $\critic: M \rightarrow \mathbb{R}$ {acts as a distinct evaluator for a specific pharmaceutical attribute}.
We built the composite score reward function as follow:
\begin{align} \label{eq:reward1}
{\RFunc^{\textsc{composite}}\paren{M}}:=
\sum_{i=1}^{|\mathbf{C}|}\text{Norm}\paren{\critic_i{\paren{M}}},\notag
\end{align}
where Norm is employed to standardize diverse attributes to a consistent scale. 
For Pareto optimality, we identify the Pareto front by iteratively collected fronts $\mathcal{P}_1, \mathcal{P}_2, \cdots, \mathcal{P}_n$ until the replay buffer $\RationaleBuffer$ is empty. These fronts partitions $\RationaleBuffer$ with $\mathcal{P}_1 \cap \mathcal{P}_2=\emptyset$ for $i\neq j$ and $\RationaleBuffer=\cup_i \mathcal{P}_i$. The ranking method assign each point $M\in \mathcal{P}_j$ the rank same as the reward as
\begin{align}
\RFunc^{\textsc{Pareto}}\paren{M}= \textsc{rank}_{\RationaleBuffer}\paren{M}=\sum_i|\mathcal{P}_i|, \forall M \in \mathcal{P}_i.
\end{align}
Both models are used to evaluate each collected molecule in the replay buffer $\RationaleBuffer$. We then reconstruct our offline dataset \dataset
\begin{align}
\dataset \leftarrow 80\% \cdot \textsc{Priority}\paren{\RationaleBuffer,\RFunc^{\textsc{Pareto}},\RFunc^{\textsc{Composite}}}~\cup~ 20\% \cdot \textsc{Random}\paren{\dataset}    
\end{align}
by randomly sampling 20\% from the previous $\dataset$ and 80\% from the buffer, with sampling prioritized according to their ranking~\citep{liu2025active}. 
We then proceed to Stage 1, where we retrain the learner policy on the dataset aligned with Pareto optimality and composite score.

\section{EXPERIMENTS}

\subsection{Experimental configuration}\label{exp_config}

\textbf{The language model.} We use GPT-2-like Transformers for causal language modeling. The training process is structured into three phases: pretraining, fine-tuning, as outlined in Algorithm \algname \secref{alg}

\fix{
\textbf{Baselines.} We benchmark our approach against baseline model, Link-INVENT \citep{guo2023linkinvent}, which employs an encoder–decoder architecture based on recurrent neural networks (RNNs). In this framework, two molecular fragments, referred to as \textit{warheads}, are provided as input constraints to the RNN-based generator, which then produces a suitable linker to connect them.
}

\textbf{Dataset.}
We utilize 1 million compounds from the ZINC15 database, docked against the human cancer protein RTCB (PDB ID: 4DWQ), as provided in the most recent Cancer dataset \citep{liu2023drugimprover, liu2025drugimprovergpt}.

\textbf{Critics and evaluation metric.} 
In this work, we assess the effectiveness of \algname in generating molecules that exhibit favorable characteristics for pharmaceutical drug discovery. To perform this evaluation, we employ the RDKit chemoinformatics toolkit \citep{landrum2016rdkit} and consider the following performance metrics.  
\textbf{Validity} determines whether the generated SMILES strings are syntactically correct.  
\textbf{Druglikeness} evaluates the probability that a molecule could serve as a viable drug candidate.  
\textbf{Solubility} estimates the likelihood that a molecule can dissolve in water, approximated by the water–octanol partition coefficient (LogP).  
\textbf{Synthesizability} measures the practical feasibility of chemical synthesis, where a score of 1 indicates high ease of synthesis and 10 indicates high difficulty \citep{ertl2009estimation}.  
\textbf{Docking Score} reflects the binding potential of a molecule to the target site; for computational efficiency, this score is predicted using a surrogate docking model (see \appref{app:surrogate_model}).  
\textbf{Similarity} is quantified by the Tanimoto coefficient, comparing the structural similarity between the generated SMILES and the original molecules.  
\textbf{Average Top 10\% Normalized Reward} captures the mean normalized reward across the top 10\% of molecules, ranked by their normalized reward.  
\textbf{Average Normalized Reward}, our primary evaluation metric, represents the mean normalized values of docking score, druglikeness, synthesizability, solubility, and similarity over all valid molecules. All evaluations are conducted on a sample of 1,280 molecules.

\subsection{Main results}

\begin{table*}[t!]
\setlength{\tabcolsep}{4pt}
   \centering
    {\small
    \scalebox{0.6}{
    \begin{tabular}{l l c c c c c c c c c c }
        \toprule
        \textbf{Target} %
        & \textbf{Algorithm}
        & {\makecell[c]{Validity~$\uparrow$}}
        & {\makecell[c]{Avg \\ Norm Reward~$\uparrow$$^{{\star}}$}}
        & {\makecell[c]{Avg Top 10 \% \\ Norm Reward~$\uparrow$}}
        & {\makecell[c]{Docking ~$\downarrow$}}
        & {\makecell[c]{Druglikeliness ~$\uparrow$}}
        & {\makecell[c]{Synthesizability ~$\downarrow$}}
        & {\makecell[c]{Solubility ~$\uparrow$}}
        & {\makecell[c]{Similarity~$\uparrow$}}
        
        \\
        \midrule
        \makecell[l]{} %
        &  \textbf{\makecell[l]{Original}}
        &  \makecell[l]{-}
        &  \makecell[l]{0.540}
        &  \makecell[l]{0.768}
        &  \makecell[l]{-8.505}
        &  \makecell[l]{0.720}
        &  \makecell[l]{2.981}
        &  \makecell[l]{2.364}
        &  \makecell[l]{-}
        \\
        \bottomrule

        \textbf{Link}
        &  \textbf{\makecell[l]{Link-INVENT \citep{guo2023linkinvent}}}
        &  \makecell[l]{0.517}
        &  \makecell[l]{0.515}
        &  \makecell[l]{0.609}
        &  \makecell[l]{-7.379}
        &  \makecell[l]{0.540}
        &  \makecell[l]{3.059}
        &  \makecell[l]{4.003}
        &  \makecell[l]{0.521}
        \\
        \textbf{ }
        &  \textbf{\makecell[l]{\algname (Pretrained)}}
        &  \makecell[l]{0.979}
        &  \makecell[l]{0.563}
        &  \makecell[l]{0.738}
        &  \makecell[l]{-8.206}
        &  \makecell[l]{0.701}
        &  \makecell[l]{2.815}
        &  \makecell[l]{2.920}
        &  \makecell[l]{0.590}
        \\
        \textbf{ }
        &  \textbf{\makecell[l]{\algname (Finetuned)}}
        &  \makecell[l]{0.991}
        &  \makecell[l]{0.616}
        &  \makecell[l]{0.734}
        &  \makecell[l]{-8.679}
        &  \makecell[l]{0.730}
        &  \makecell[l]{2.418}
        &  \makecell[l]{3.684}
        &  \makecell[l]{0.564}
        \\
        \bottomrule
        \textbf{Merge}
        &  \textbf{\makecell[l]{Link-INVENT \citep{guo2023linkinvent}}}
        &  \makecell[l]{0.409}
        &  \makecell[l]{0.443}
        &  \makecell[l]{0.510}
        &  \makecell[l]{-2.529}
        &  \makecell[l]{0.229}
        &  \makecell[l]{3.607}
        &  \makecell[l]{5.257}
        &  \makecell[l]{0.720}
        \\
        \textbf{ }
        &  \textbf{\makecell[l]{\algname (Pretrained)}}
        &  \makecell[l]{0.947}
        &  \makecell[l]{0.630}
        &  \makecell[l]{0.788}
        &  \makecell[l]{-8.235}
        &  \makecell[l]{0.692}
        &  \makecell[l]{2.859}
        &  \makecell[l]{2.829}
        &  \makecell[l]{0.953}
        \\
        \textbf{ }
        &  \textbf{\makecell[l]{\algname (Finetuned)}}
        &  \makecell[l]{ 0.993}
        &  \makecell[l]{0.651}
        &  \makecell[l]{0.791}
        &  \makecell[l]{-8.375}
        &  \makecell[l]{0.727}
        &  \makecell[l]{2.699}
        &  \makecell[l]{2.815}
        &  \makecell[l]{0.965}
        \\
        \bottomrule
        \\
        \end{tabular}}}
        \caption{
        {\textbf{Main results.} A comparison of Link-INVENT and different versions of \algname on various objectives  
        based on RTCB (PDBID: 4DWQ) datasets. 
        }
        }
        \label{exp:main_result}
\end{table*}

Table \ref{exp:main_result} demonstrates that \algname substantially outperforms Link-INVENT across both linking and merging tasks. This advantage arises from both architectural and methodological differences: \algname employs a transformer-based GPT backbone, which is more expressive than the RNN architecture of Link-INVENT, and it explicitly incorporates common substructures when merging fragments. In contrast, Link-INVENT neglects common substructures, often producing unnecessarily complex molecules when handling fragments with overlapping motifs, which leads to degraded performance. Moreover, the finetuned version of \algname consistently surpasses the pretrained model across nearly all metrics, highlighting the effectiveness of our proposed RAE algorithm in improving molecular generation.

\subsubsection{Fragment Growing}\label{app:growing}
\begin{figure}[ht]
    \centering
    \includegraphics[width=1\linewidth]{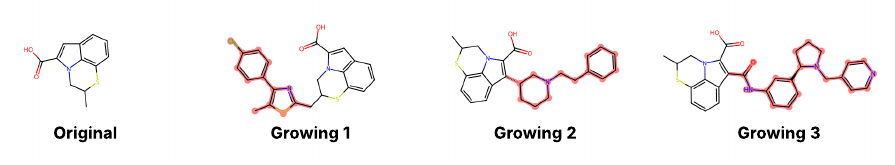}
    \caption{Given an initial fragment, \algname can effectively expand and refine it into a more complete molecular structure. The \textcolor{red}{highlighted substructures} indicate the newly grown parts.}
    \label{fig:growing_gen_example}
\end{figure}

\fix{Figure \ref{fig:growing_gen_example} illustrates examples of fragment growing, where the original molecule was obtained from an NMR-based screening \citep{g2017protein}. This example highlights the ability of \algname to effectively transform simple fragments into chemically plausible and drug-like molecules.}

\begin{figure}[ht]
    \centering
    \includegraphics[width=0.25\linewidth]{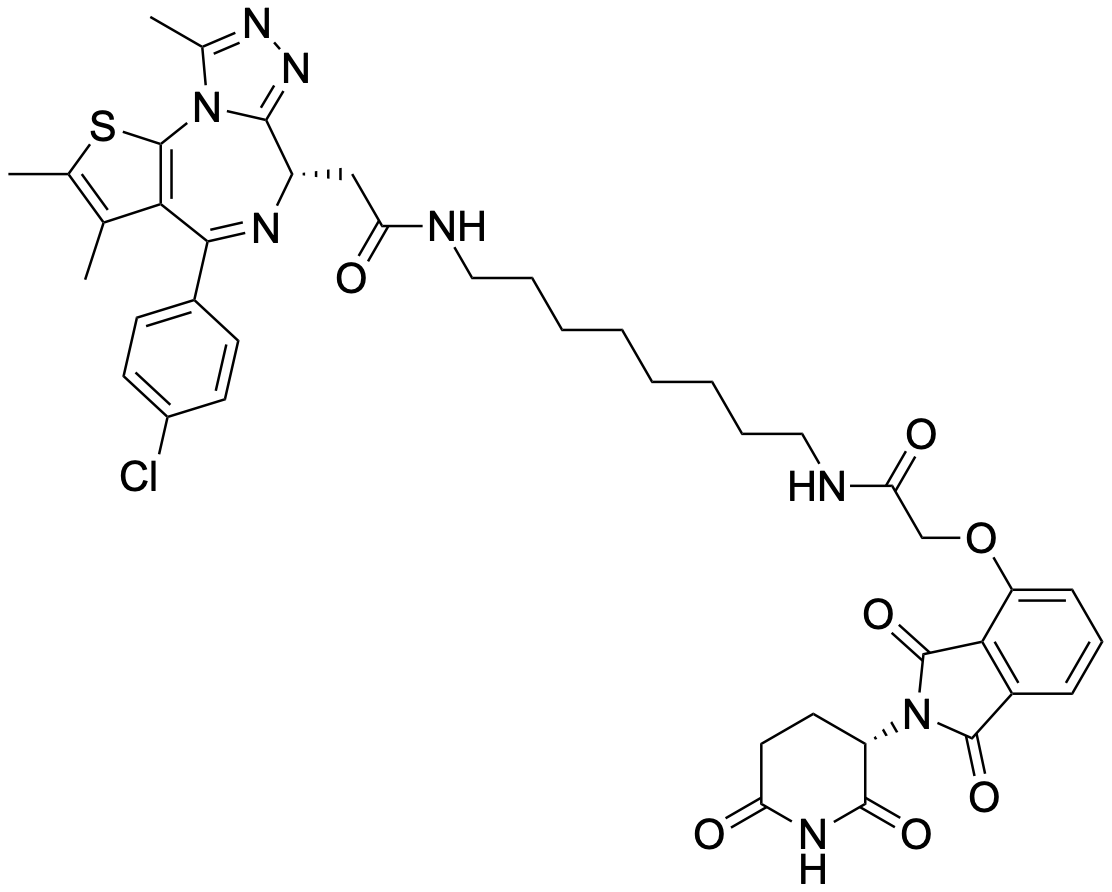}
    \caption{dBET6 Chemical Structure}
    \label{fig:dBET6_viz}
\end{figure}

\begin{figure}[ht]
    \centering
    \includegraphics[width=0.9\linewidth]{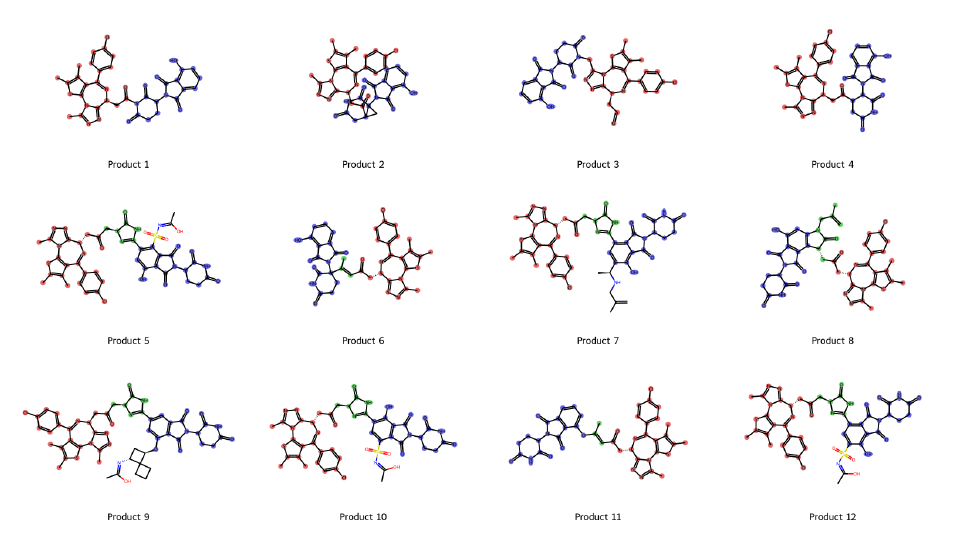}
    \caption{Chemical structure of a bifunctional PROTAC degrader. The molecule is designed to induce proximity between an E3 ubiquitin ligase and a Protein of Interest (POI). The E3-recruiting fragment is colored \textcolor{red}{red}, the POI-binding warhead is \textcolor{blue}{blue}, and the interconnecting linker is highlighted in \textcolor{green}{green}.}
    \label{fig:linker_example}
\end{figure}

\subsubsection{Fragment Linking}\label{app:linking}

dBET6, as visualized in figure \ref{fig:dBET6_viz}, is a PROTAC molecule with high structural selectivity that bridges the E3 ubiquitin ligase CRBN and the target protein BRD4 \citep{winter2017bet}. We use this example to investigate whether our model can identify reasonable linkage sites and propose a suitable linker, and results are represented in figure \ref{fig:linker_example}.

The molecular generation process predominantly yielded structures where the two fragments are connected via a linker, with only a few exceptions showing direct linking. All generated linkers were determined to be chemically stable. Notably, several of these linkers incorporate an imidazolinone scaffold. This structural motif is highly significant in medicinal chemistry, widely utilized in drug molecules for its chemical stability and biocompatibility. Furthermore, a particularly noteworthy achievement is that our model successfully identified the acyl linking site on one of the fragments, despite not being provided with any explicit information about this site in the inference stage. This demonstrates the model's robust capability for implicit learning and recognition of key chemical features.

\subsubsection{Fragment Merging}\label{app:merging}

\begin{figure}[ht]
    \centering
    \includegraphics[width=0.9\linewidth]{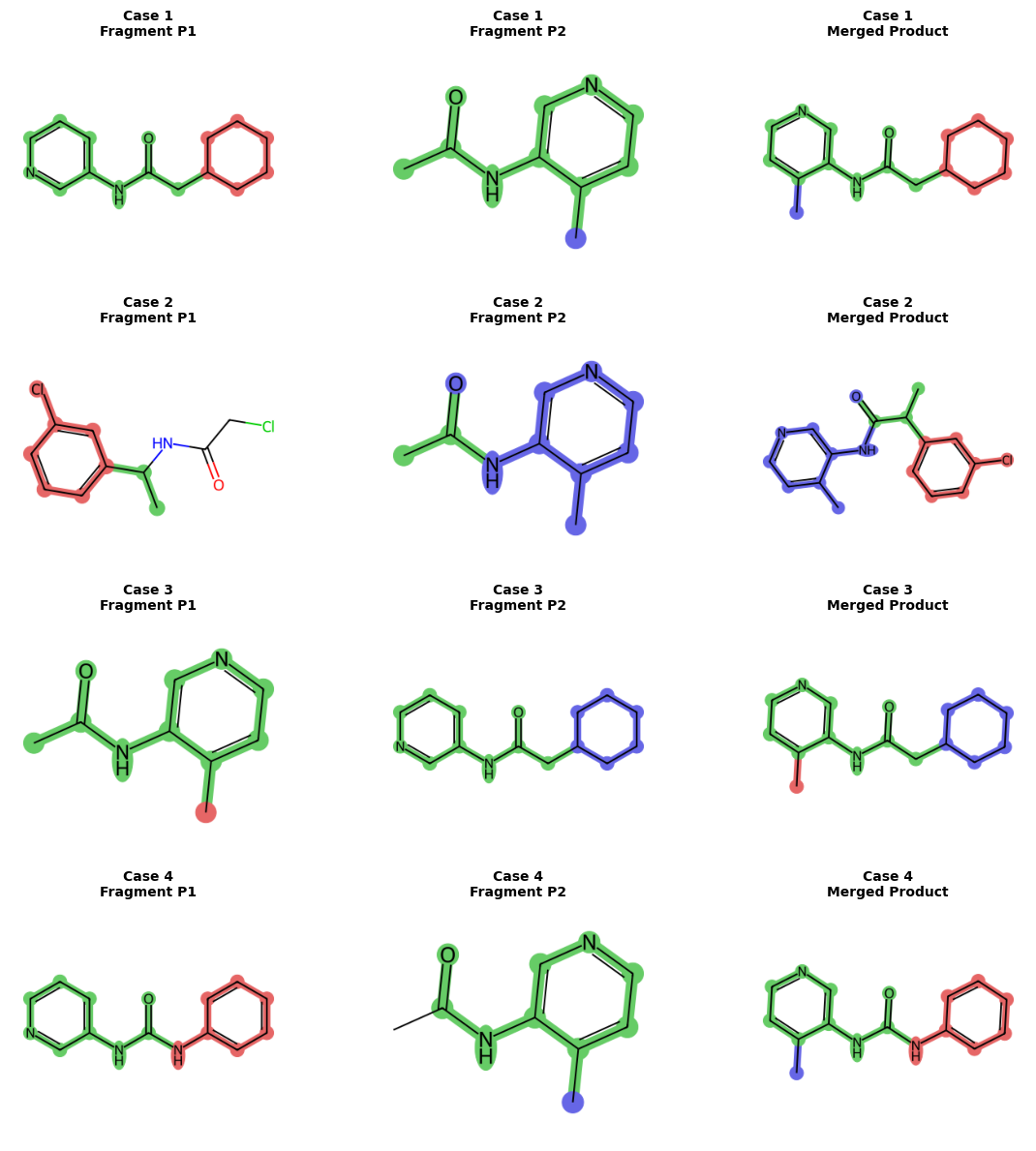}
    \caption{Four merging cases were generated, with all generated molecules contained within the 24 molecules of submission TRY-UNI-714a760b. Different fragments are marked in \textcolor{red}{red} and \textcolor{blue}{blue}, while the merged overlapping regions are highlighted in \textcolor{green}{green}.}
    \label{fig:merging_example}
\end{figure}

To assess our model's ability to merge different fragments, we utilized the COVID Moonshot dataset \citep{boby2023open}—a fully open science, crowd-sourced, and structure-enabled drug discovery campaign against the severe acute respiratory syndrome coronavirus 2 (SARS-CoV-2) main protease (Mpro). This dataset provides an ideal testbed for evaluating fragment-based drug design approaches due to its comprehensive collection of crystallographic fragment hits and subsequent rational design submissions.
The submission TRY-UNI-714a760b represents a class of rationally designed molecules that were constructed based on the strategic combination of five distinct fragments identified through crystallographic screening. These fragments were carefully selected and merged by human medicinal chemists who recognized their spatial complementarity and potential for synergistic binding interactions within the Mpro active site.
To evaluate our model's fragment merging capabilities, we challenged it to reproduce this human-designed approach by providing the same five fragments as input and tasking it with generating novel molecular combinations. Remarkably, our model successfully generated four molecules among the twenty four molecules present in the submission series, as shown in figure \ref{fig:merging_example}. Our model can efficiently identify a variety of drug-related functional groups, including pyridine rings, amide bonds, and others. This achievement demonstrates our model's sophisticated ability to identify mergeable sites and recognize the structural compatibility between different fragment scaffolds.

The success in reproducing four out of five fragments highlights several key strengths of our approach: (1) the model's capacity to recognize pharmacophoric features that are essential for maintaining binding affinity, (2) its ability to identify chemically feasible linking strategies between fragments. This performance validates our model's potential as a valuable tool for fragment-based drug design, particularly in its ability to automate and accelerate the traditionally labor-intensive process of fragment merging and optimization.

\section{CONCLUSION}

\fix{
We have introduced \algname, a novel GPT-based framework for fragment-based drug discovery that integrates chemically-aware, energy-based bond cleavage pre-training. This design equips the model with fragment growing, linking, and merging capabilities. In addition, we proposed Reward Ranked Alignment with Expert Exploration (RAE), a SFT alignment strategy that leverages expert imitation for diversity, data selection and augmentation for Pareto and composite score optimization, and supervised fine-tuning for aligning policy behavior with multi-objective drug design goals.
Through comprehensive evaluations on real-world cancer-related datasets, \algname consistently outperformed competitive baselines across a wide range of metrics, underscoring its effectiveness in generating drug-like molecules with improved pharmaceutical properties. 
While the current implementation is limited to SMILES-based representations, we are working to extend \algname to broader molecular formats and structural representations.
}

\subsubsection*{Acknowledgements}
This work is supported in part by the RadBio-AI project (DE-AC02-06CH11357), U.S. Department of Energy Office of Science, Office of Biological and Environment Research, the Improve project under contract (75N91019F00134, 75N91019D00024, 89233218CNA000001, DE-AC02-06-CH11357, DE-AC52-07NA27344, DE-AC05-00OR22725), 
the Exascale Computing Project (17-SC-20-SC), a collaborative effort of the U.S. Department of Energy Office of Science and the National Nuclear Security Administration.

\bibliographystyle{plainnat}
\bibliography{reference}

\begin{thebibliography}{41}
\providecommand{\natexlab}[1]{#1}
\providecommand{\url}[1]{\texttt{#1}}
\expandafter\ifx\csname urlstyle\endcsname\relax
  \providecommand{\doi}[1]{doi: #1}\else
  \providecommand{\doi}{doi: \begingroup \urlstyle{rm}\Url}\fi

\bibitem[Abeer et~al.(2024)Abeer, Urban, Weil, Alexander, and Yoon]{abeer2024multi}
ANM~Nafiz Abeer, Nathan~M Urban, M~Ryan Weil, Francis~J Alexander, and Byung-Jun Yoon.
\newblock Multi-objective latent space optimization of generative molecular design models.
\newblock \emph{Patterns}, 5\penalty0 (10), 2024.

\bibitem[Addie et~al.(2013)Addie, Ballard, Buttar, Crafter, Currie, Davies, Debreczeni, Dry, Dudley, Greenwood, Johnson, Kettle, Lane, Lamont, Leach, Luke, Morris, Ogilvie, Page, Pass, Pearson, and Ruston]{addie2013astra}
Matt Addie, Peter Ballard, David Buttar, Claire Crafter, Gordon Currie, Barry~R. Davies, Judit Debreczeni, Hannah Dry, Philippa Dudley, Ryan Greenwood, Paul~D. Johnson, Jason~G. Kettle, Clare Lane, Gillian Lamont, Andrew Leach, Richard W.~A. Luke, Jeff Morris, Donald Ogilvie, Ken Page, Martin Pass, Stuart Pearson, and Linette Ruston.
\newblock Discovery of 4-amino-n-[(1s)-1-(4-chlorophenyl)-3-hydroxypropyl]-1-(7h-pyrrolo[2,3-d]pyrimidin-4-yl)piperidine-4-carboxamide (azd5363), an orally bioavailable, potent inhibitor of akt kinases.
\newblock \emph{Journal of Medicinal Chemistry}, 56\penalty0 (5):\penalty0 2059--2073, 2013.
\newblock \doi{10.1021/jm301762v}.
\newblock URL \url{https://doi.org/10.1021/jm301762v}.
\newblock PMID: 23394218.

\bibitem[Boby et~al.(2023)Boby, Fearon, Ferla, Filep, Koekemoer, Robinson, Consortium‡, Chodera, Lee, London, et~al.]{boby2023open}
Melissa~L Boby, Daren Fearon, Matteo Ferla, Mihajlo Filep, Lizb{\'e} Koekemoer, Matthew~C Robinson, COVID~Moonshot Consortium‡, John~D Chodera, Alpha~A Lee, Nir London, et~al.
\newblock Open science discovery of potent noncovalent sars-cov-2 main protease inhibitors.
\newblock \emph{Science}, 382\penalty0 (6671):\penalty0 eabo7201, 2023.

\bibitem[Bollag et~al.(2012)Bollag, Tsai, Zhang, Zhang, Ibrahim, Nolop, and Hirth]{bollag2012vemurafenib}
Gideon Bollag, Jack Tsai, Jian Zhang, Christina Zhang, Pamela Ibrahim, Keith Nolop, and Paul Hirth.
\newblock Vemurafenib: the first drug approved for braf-mutant cancer.
\newblock \emph{Nature Reviews Drug Discovery}, 11\penalty0 (11):\penalty0 873--886, Nov 2012.
\newblock \doi{10.1038/nrd3847}.
\newblock URL \url{https://doi.org/10.1038/nrd3847}.
\newblock Epub 2012 Oct 12. PMID: 23060265.

\bibitem[Ertl and Schuffenhauer(2009)]{ertl2009estimation}
Peter Ertl and Ansgar Schuffenhauer.
\newblock Estimation of synthetic accessibility score of drug-like molecules based on molecular complexity and fragment contributions.
\newblock \emph{Journal of cheminformatics}, 1:\penalty0 1--11, 2009.

\bibitem[Exertier et~al.(2024)Exertier, Salerno, Antonelli, Fiorillo, Ocello, Seghetti, Caciolla, Uliassi, Masetti, Fiorentino, Orsini, Di~Muccio, Ilari, and Bolognesi]{Exertier2024}
C{\'e}cile Exertier, Alessandra Salerno, Lorenzo Antonelli, Annarita Fiorillo, Riccardo Ocello, Francesca Seghetti, Jessica Caciolla, Elisa Uliassi, Matteo Masetti, Eleonora Fiorentino, Stefania Orsini, Trentina Di~Muccio, Andrea Ilari, and Maria~Laura Bolognesi.
\newblock Fragment merging, growing, and linking identify new trypanothione reductase inhibitors for leishmaniasis.
\newblock \emph{Journal of Medicinal Chemistry}, 67\penalty0 (1):\penalty0 402--419, 2024.
\newblock \doi{10.1021/acs.jmedchem.3c01439}.
\newblock URL \url{https://doi.org/10.1021/acs.jmedchem.3c01439}.
\newblock PMID: 38164929.

\bibitem[Fairbrother et~al.(2019)Fairbrother, Leverson, Sampath, and Souers]{fairbrother2019venetoclax}
W.~J. Fairbrother, J.~D. Leverson, D.~Sampath, and A.~J. Souers.
\newblock Discovery and development of venetoclax, a selective antagonist of bcl‐2.
\newblock In J.~Fischer, C.~Klein, and W.~E. Childers, editors, \emph{Successful Drug Discovery}, volume~4, pages 225--245. Wiley‐VCH, Weinheim, 2019.
\newblock [Google Scholar].

\bibitem[G.~Ferreira and D.~Andricopulo(2017)]{g2017protein}
Leonardo G.~Ferreira and Adriano D.~Andricopulo.
\newblock From protein structure to small-molecules: recent advances and applications to fragment-based drug discovery.
\newblock \emph{Current topics in medicinal chemistry}, 17\penalty0 (20):\penalty0 2260--2270, 2017.

\bibitem[Gage(1994)]{gage1994new_bpe}
Philip Gage.
\newblock A new algorithm for data compression.
\newblock \emph{The C Users Journal}, 12\penalty0 (2):\penalty0 23--38, 1994.

\bibitem[Guan et~al.(2023)Guan, Peng, Jiang, Luo, Peng, and Ma]{guan2023linkernet}
Jiaqi Guan, Xingang Peng, Peiqi Jiang, Yunan Luo, Jian Peng, and Jianzhu Ma.
\newblock Linkernet: fragment poses and linker co-design with 3d equivariant diffusion.
\newblock In \emph{Proceedings of the 37th International Conference on Neural Information Processing Systems}, NIPS '23, Red Hook, NY, USA, 2023. Curran Associates Inc.

\bibitem[Guo et~al.(2023)Guo, Knuth, Margreitter, Janet, Papadopoulos, Engkvist, and Patronov]{guo2023linkinvent}
Jeff Guo, Franziska Knuth, Christian Margreitter, Jon~Paul Janet, Kostas Papadopoulos, Ola Engkvist, and Atanas Patronov.
\newblock Link-invent: generative linker design with reinforcement learning.
\newblock \emph{Digital Discovery}, 2\penalty0 (2):\penalty0 392–408, April 2023.
\newblock ISSN 2635-098X.
\newblock \doi{10.1039/D2DD00115B}.
\newblock URL \url{https://pubs.rsc.org/en/content/articlelanding/2023/dd/d2dd00115b}.

\bibitem[Heinrich et~al.(2013)Heinrich, Seenisamy, Emmanuvel, Kulkarni, Bomke, Rohdich, Greiner, Esdar, Krier, Gradler, et~al.]{heinrich2013fragment}
Timo Heinrich, Jeyaprakashnarayanan Seenisamy, Lourdusamy Emmanuvel, Santosh Kulkarni, Jorg Bomke, Felix Rohdich, Hartmut Greiner, Christina Esdar, Mireille Krier, Ulrich Gradler, et~al.
\newblock Fragment-based discovery of new highly substituted 1 h-pyrrolo [2, 3-b]-and 3 h-imidazolo [4, 5-b]-pyridines as focal adhesion kinase inhibitors.
\newblock \emph{Journal of medicinal chemistry}, 56\penalty0 (3):\penalty0 1160--1170, 2013.

\bibitem[Huang et~al.(2022)Huang, Peng, Ma, and Zhang]{huang20223dlinker}
Yinan Huang, Xingang Peng, Jianzhu Ma, and Muhan Zhang.
\newblock 3{DL}inker: An e(3) equivariant variational autoencoder for molecular linker design.
\newblock In Kamalika Chaudhuri, Stefanie Jegelka, Le~Song, Csaba Szepesvari, Gang Niu, and Sivan Sabato, editors, \emph{Proceedings of the 39th International Conference on Machine Learning}, volume 162 of \emph{Proceedings of Machine Learning Research}, pages 9280--9294. PMLR, 17--23 Jul 2022.
\newblock URL \url{https://proceedings.mlr.press/v162/huang22g.html}.

\bibitem[Igashov et~al.(2024)Igashov, Stärk, Vignac, Schneuing, Satorras, Frossard, Welling, Bronstein, and Correia]{igashov2024difflinker}
Ilia Igashov, Hannes Stärk, Clément Vignac, Arne Schneuing, Victor~Garcia Satorras, Pascal Frossard, Max Welling, Michael Bronstein, and Bruno Correia.
\newblock Equivariant 3d-conditional diffusion model for molecular linker design.
\newblock \emph{Nature Machine Intelligence}, 6\penalty0 (4):\penalty0 417–427, April 2024.
\newblock ISSN 2522-5839.
\newblock \doi{10.1038/s42256-024-00815-9}.
\newblock URL \url{https://www.nature.com/articles/s42256-024-00815-9}.

\bibitem[Keserű et~al.(2016)Keserű, Erlanson, Ferenczy, Hann, Murray, and Pickett]{Keserű2016}
Gy{\"o}rgy~M. Keserű, Daniel~A. Erlanson, Gy{\"o}rgy~G. Ferenczy, Michael~M. Hann, Christopher~W. Murray, and Stephen~D. Pickett.
\newblock Design principles for fragment libraries: Maximizing the value of learnings from pharma fragment-based drug discovery (fbdd) programs for use in academia.
\newblock \emph{Journal of Medicinal Chemistry}, 59\penalty0 (18):\penalty0 8189--8206, 2016.
\newblock \doi{10.1021/acs.jmedchem.6b00197}.
\newblock URL \url{https://doi.org/10.1021/acs.jmedchem.6b00197}.
\newblock PMID: 27124799.

\bibitem[Landrum et~al.()]{landrum2016rdkit}
Greg Landrum et~al.
\newblock R{D}kit: Open-source cheminformatics software.
\newblock \url{https://www.rdkit.org}. Accessed Oct 2023.

\bibitem[Lanman et~al.(2020)Lanman, Allen, Allen, Amegadzie, Ashton, Booker, Chen, Chen, Frohn, Goodman, Kopecky, Liu, Lopez, Low, Ma, Minatti, Nguyen, Nishimura, Pickrell, Reed, Shin, Siegmund, Tamayo, Tegley, Walton, Wang, Wurz, Xue, Yang, Achanta, Bartberger, Canon, Hollis, McCarter, Mohr, Rex, Saiki, San~Miguel, Volak, Wang, Whittington, Zech, Lipford, and Cee]{Lanman2020Sotorasib}
Brian~A. Lanman, Jennifer~R. Allen, John~G. Allen, Albert~K. Amegadzie, Kate~S. Ashton, Shon~K. Booker, Jian~Jeffrey Chen, Ning Chen, Michael~J. Frohn, Guy Goodman, David~J. Kopecky, Longbin Liu, Patricia Lopez, Jonathan~D. Low, Vu~Ma, Ana~E. Minatti, Thomas~T. Nguyen, Nobuko Nishimura, Alexander~J. Pickrell, Anthony~B. Reed, Youngsook Shin, Aaron~C. Siegmund, Nuria~A. Tamayo, Christopher~M. Tegley, Mary~C. Walton, Hui-Ling Wang, Ryan~P. Wurz, May Xue, Kevin~C. Yang, Pragathi Achanta, Michael~D. Bartberger, Jude Canon, L.~Steven Hollis, John~D. McCarter, Christopher Mohr, Karen Rex, Anne~Y. Saiki, Tisha San~Miguel, Laurie~P. Volak, Kevin~H. Wang, Douglas~A. Whittington, Stephan~G. Zech, J.~Russell Lipford, and Victor~J. Cee.
\newblock Discovery of a covalent inhibitor of krasg12c (amg 510) for the treatment of solid tumors.
\newblock \emph{Journal of Medicinal Chemistry}, 63\penalty0 (1):\penalty0 52--65, 2020.
\newblock \doi{10.1021/acs.jmedchem.9b01180}.
\newblock URL \url{https://doi.org/10.1021/acs.jmedchem.9b01180}.
\newblock PMID: 31820981.

\bibitem[Leach and Hann(2011)]{LEACH2011489}
Andrew~R Leach and Michael~M Hann.
\newblock Molecular complexity and fragment-based drug discovery: ten years on.
\newblock \emph{Current Opinion in Chemical Biology}, 15\penalty0 (4):\penalty0 489--496, 2011.
\newblock ISSN 1367-5931.
\newblock \doi{https://doi.org/10.1016/j.cbpa.2011.05.008}.
\newblock URL \url{https://www.sciencedirect.com/science/article/pii/S1367593111000718}.
\newblock Next Generation Therapeutics.

\bibitem[Li et~al.(2024)Li, Hu, Zhou, Yang, and Bai]{li2024diffprotacs}
Fenglei Li, Qiaoyu Hu, Yongqi Zhou, Hao Yang, and Fang Bai.
\newblock Diffprotacs is a deep learning-based generator for proteolysis targeting chimeras.
\newblock \emph{Briefings in Bioinformatics}, 25\penalty0 (5):\penalty0 bbae358, 08 2024.
\newblock ISSN 1477-4054.
\newblock \doi{10.1093/bib/bbae358}.
\newblock URL \url{https://doi.org/10.1093/bib/bbae358}.

\bibitem[Liu et~al.({\natexlab{a}})Liu, Jiang, Foster, Xu, and Stevens]{liu2025scaffold}
Xuefeng Liu, Songhao Jiang, Ian Foster, Jinbo Xu, and Rick~L Stevens.
\newblock Scaffold-driven gpt model for drug optimization.
\newblock In \emph{ICML 2025 Generative AI and Biology (GenBio) Workshop}, {\natexlab{a}}.

\bibitem[Liu et~al.({\natexlab{b}})Liu, Jiang, Li, and Stevens]{liu2025ground}
Xuefeng Liu, Songhao Jiang, Bo~Li, and Rick~L Stevens.
\newblock A ground-up designed controllable gpt for molecule optimization.
\newblock In \emph{ICML 2025 Generative AI and Biology (GenBio) Workshop}, {\natexlab{b}}.

\bibitem[Liu et~al.(2022)Liu, Xia, Stevens, and Chen]{liu2022cost}
Xuefeng Liu, Fangfang Xia, Rick~L Stevens, and Yuxin Chen.
\newblock Cost-effective online contextual model selection.
\newblock \emph{arXiv preprint arXiv:2207.06030}, 2022.

\bibitem[Liu et~al.(2023{\natexlab{a}})Liu, Jiang, Vasan, Brace, Gokdemir, Brettin, Xia, Foster, and Stevens]{liu2023drugimprover}
Xuefeng Liu, Songhao Jiang, Archit Vasan, Alexander Brace, Ozan Gokdemir, Thomas Brettin, Fangfang Xia, Ian Foster, and Rick Stevens.
\newblock Drugimprover: Utilizing reinforcement learning for multi-objective alignment in drug optimization.
\newblock In \emph{NeurIPS 2023 Workshop on New Frontiers of AI for Drug Discovery and Development}, 2023{\natexlab{a}}.

\bibitem[Liu et~al.(2023{\natexlab{b}})Liu, Yoneda, Stevens, Walter, and Chen]{liu2023blending}
Xuefeng Liu, Takuma Yoneda, Rick~L Stevens, Matthew~R Walter, and Yuxin Chen.
\newblock Blending imitation and reinforcement learning for robust policy improvement.
\newblock \emph{arXiv preprint arXiv:2310.01737}, 2023{\natexlab{b}}.

\bibitem[Liu et~al.(2025{\natexlab{a}})Liu, Jiang, Chen, Yang, Chen, Foster, and Stevens]{liu2025drugimprovergpt}
Xuefeng Liu, Songhao Jiang, Siyu Chen, Zhuoran Yang, Yuxin Chen, Ian Foster, and Rick Stevens.
\newblock Drugimprovergpt: A large language model for drug optimization with fine-tuning via structured policy optimization.
\newblock \emph{arXiv preprint arXiv:2502.07237}, 2025{\natexlab{a}}.

\bibitem[Liu et~al.(2025{\natexlab{b}})Liu, Jiang, Foster, Xu, and Stevens]{liu2025scaffoldgpt}
Xuefeng Liu, Songhao Jiang, Ian Foster, Jinbo Xu, and Rick Stevens.
\newblock Scaffoldgpt: A scaffold-based gpt model for drug optimization.
\newblock \emph{arXiv preprint arXiv:2502.06891}, 2025{\natexlab{b}}.

\bibitem[Liu et~al.(2025{\natexlab{c}})Liu, Jiang, Li, and Stevens]{liu2025controllablegpt}
Xuefeng Liu, Songhao Jiang, Bo~Li, and Rick Stevens.
\newblock Controllablegpt: A ground-up designed controllable gpt for molecule optimization.
\newblock \emph{arXiv preprint arXiv:2502.10631}, 2025{\natexlab{c}}.

\bibitem[Liu et~al.(2025{\natexlab{d}})Liu, Le, Chen, Stevens, Yang, Walter, and Chen]{liu2025active}
Xuefeng Liu, Hung~TC Le, Siyu Chen, Rick Stevens, Zhuoran Yang, Matthew~R Walter, and Yuxin Chen.
\newblock Active advantage-aligned online reinforcement learning with offline data.
\newblock \emph{arXiv preprint arXiv:2502.07937}, 2025{\natexlab{d}}.

\bibitem[Murray and Rees(2009)]{murray2009rise}
Christopher~W. Murray and David~C. Rees.
\newblock The rise of fragment-based drug discovery.
\newblock \emph{Nature Chemistry}, 1\penalty0 (3):\penalty0 187--192, 2009.
\newblock \doi{10.1038/nchem.217}.
\newblock URL \url{https://doi.org/10.1038/nchem.217}.

\bibitem[Ren et~al.(2014)Ren, Li, Wang, Chen, Shen, Liu, Chen, Cao, Li, Zhang, Xu, Geng, He, Xiong, and Shen]{Ren2014HSP90}
Jie Ren, Jie Li, Yu~Wang, Wei Chen, An~Shen, Honglin Liu, Dan Chen, Dongmei Cao, Ying Li, Nan Zhang, Yili Xu, Meiyu Geng, Jian He, Bing Xiong, and Jianping Shen.
\newblock Identification of a new series of potent diphenol hsp90 inhibitors by fragment merging and structure-based optimization.
\newblock \emph{Bioorganic \& Medicinal Chemistry Letters}, 24\penalty0 (11):\penalty0 2525--2529, June 2014.
\newblock ISSN 0960-894X.
\newblock \doi{10.1016/j.bmcl.2014.03.100}.
\newblock URL \url{https://doi.org/10.1016/j.bmcl.2014.03.100}.
\newblock Epub 2014 Apr 8.

\bibitem[Sennrich et~al.(2015)Sennrich, Haddow, and Birch]{sennrich2015neural_bpe}
Rico Sennrich, Barry Haddow, and Alexandra Birch.
\newblock Neural machine translation of rare words with subword units.
\newblock \emph{arXiv preprint arXiv:1508.07909}, 2015.

\bibitem[Smith et~al.(2022)Smith, Aranda, Bobinski, Briere, Burns, Christensen, Clarine, Engstrom, Gunn, Ivetac, et~al.]{smith2022fragment}
Christopher~R Smith, Ruth Aranda, Thomas~P Bobinski, David~M Briere, Aaron~C Burns, James~G Christensen, Jeffery Clarine, Lars~D Engstrom, Robin~J Gunn, Anthony Ivetac, et~al.
\newblock Fragment-based discovery of mrtx1719, a synthetic lethal inhibitor of the prmt5• mta complex for the treatment of mtap-deleted cancers.
\newblock \emph{Journal of Medicinal Chemistry}, 65\penalty0 (3):\penalty0 1749--1766, 2022.

\bibitem[Vasan et~al.(2023)Vasan, Stevens, Ramanathan, and Venkatram]{vasan23}
Archit Vasan, Rick Stevens, Arvind Ramanathan, and Vishwanath Venkatram.
\newblock Benchmarking language-based docking models.
\newblock 2023.

\bibitem[Wasko et~al.(2015)Wasko, Pellegrene, Madura, and Surratt]{wasko2015fragment}
Matthew~J. Wasko, Kelly~A. Pellegrene, Jeffry~D. Madura, and Christopher~K. Surratt.
\newblock A role for fragment-based drug design in developing novel lead compounds for central nervous system targets.
\newblock \emph{Frontiers in Neurology}, 6:\penalty0 197, 2015.
\newblock \doi{10.3389/fneur.2015.00197}.
\newblock URL \url{https://doi.org/10.3389/fneur.2015.00197}.

\bibitem[Weininger(1988)]{weininger1988smiles}
David Weininger.
\newblock Smiles, a chemical language and information system. 1. introduction to methodology and encoding rules.
\newblock \emph{Journal of chemical information and computer sciences}, 28\penalty0 (1):\penalty0 31--36, 1988.

\bibitem[Wills et~al.(2023)Wills, Sanchez-Garcia, Dudgeon, Roughley, Merritt, Hubbard, Davidson, von Delft, and Deane]{Wills2023}
Stephanie Wills, Ruben Sanchez-Garcia, Tim Dudgeon, Stephen~D. Roughley, Andy Merritt, Roderick~E. Hubbard, James Davidson, Frank von Delft, and Charlotte~M. Deane.
\newblock Fragment merging using a graph database samples different catalogue space than similarity search.
\newblock \emph{Journal of Chemical Information and Modeling}, 63\penalty0 (11):\penalty0 3423--3437, 2023.
\newblock \doi{10.1021/acs.jcim.3c00276}.
\newblock URL \url{https://doi.org/10.1021/acs.jcim.3c00276}.
\newblock PMID: 37229647.

\bibitem[Winter et~al.(2017)Winter, Mayer, Buckley, Erb, Roderick, Vittori, Reyes, di~Iulio, Souza, Ott, et~al.]{winter2017bet}
Georg~E Winter, Andreas Mayer, Dennis~L Buckley, Michael~A Erb, Justine~E Roderick, Sarah Vittori, Jaime~M Reyes, Julia di~Iulio, Amanda Souza, Christopher~J Ott, et~al.
\newblock Bet bromodomain proteins function as master transcription elongation factors independent of cdk9 recruitment.
\newblock \emph{Molecular cell}, 67\penalty0 (1):\penalty0 5--18, 2017.

\bibitem[Wyatt et~al.(2008)Wyatt, Woodhead, Berdini, Boulstridge, Carr, Cross, Davis, Devine, Early, Feltell, et~al.]{wyatt2008identification}
Paul~G Wyatt, Andrew~J Woodhead, Valerio Berdini, John~A Boulstridge, Maria~G Carr, David~M Cross, Deborah~J Davis, Lindsay~A Devine, Theresa~R Early, Ruth~E Feltell, et~al.
\newblock Identification of n-(4-piperidinyl)-4-(2, 6-dichlorobenzoylamino)-1 h-pyrazole-3-carboxamide (at7519), a novel cyclin dependent kinase inhibitor using fragment-based x-ray crystallography and structure based drug design.
\newblock \emph{Journal of medicinal chemistry}, 51\penalty0 (16):\penalty0 4986--4999, 2008.

\bibitem[Xu and Kang(2025)]{weijunxu2025}
Weijun Xu and Congbao Kang.
\newblock Fragment-based drug design: From then until now, and toward the future.
\newblock \emph{Journal of Medicinal Chemistry}, 68\penalty0 (5):\penalty0 5000--5004, 2025.
\newblock \doi{10.1021/acs.jmedchem.5c00424}.
\newblock URL \url{https://doi.org/10.1021/acs.jmedchem.5c00424}.
\newblock PMID: 39992814.

\bibitem[Xuefeng et~al.(2024)Xuefeng, Chih-Chan, Peng, Songhao, and Rick]{liu2024erp}
Liu Xuefeng, Tien Chih-Chan, Ding Peng, Jiang Songhao, and Stevens Rick.
\newblock Entropy-reinforced planning with large language models for de novo drug discovery.
\newblock \emph{arXiv 2024}, 2024.

\bibitem[Yang et~al.(2020)Yang, Zheng, Su, Zhao, Xu, and Chen]{yang2020syntalinker}
Yuyao Yang, Shuangjia Zheng, Shimin Su, Chao Zhao, Jun Xu, and Hongming Chen.
\newblock Syntalinker: automatic fragment linking with deep conditional transformer neural networks.
\newblock \emph{Chem. Sci.}, 11:\penalty0 8312--8322, 2020.
\newblock \doi{10.1039/D0SC03126G}.
\newblock URL \url{http://dx.doi.org/10.1039/D0SC03126G}.

\end{thebibliography}

\clearpage

\appendix

\section{Appendix}

\subsection{Pre-training and finetuning dataset}\label{app:pretrain_data}
For pretraining, we utilized the ZINC dataset, selecting only molecules categorized as Standard, In-Stock, and Drug-Like, which resulted in approximately 11 million compounds. 
Preprocessing involved two straightforward steps: (i) canonicalizing SMILES strings using \texttt{Chem.MolToSmiles(Chem.MolFromSmiles(mol), True)}, and (ii) discarding molecules with empty scaffold SMILES. These procedures are also incorporated into the HuggingFace data repository.

For finetuning, we employed 1 million compounds from the ZINC15 dataset, each docked to the RTCB protein (PDB ID: 4DWQ) implicated in human cancer, as provided by the Cancer dataset of \citet{liu2023drugimprover}, and applied them consistently across all baselines. In each finetuning epoch, a random subset of the desired number of molecules was sampled. 

For evaluation and experimentation, we used 1,280 molecules from the Cancer dataset that did not overlap with the finetuning set.

\subsection{Pretraining}

\paragraph{Incremental \fix{pre-}training.} In the first phase, we train a GPT model exclusively on molecular data using Causal Language Modeling (CLM). In this autoregressive setup, the model learns to predict the next token in a sequence conditioned only on the preceding tokens. The Phase~1 training corpus is constructed as follows:
\begin{equation}\label{eq:phase1_corpus}
\corpus_{\text{Phase 1}}=\curlybracket{[BOS],\tuple{L},\underbrace{y_1,\cdots,y_{\horizon}}_{\text{target molecule Y}},[EOS]}, \fix{\tuple{L} \text{is the token for ligand.}}
\end{equation}
In the second phase, we build upon the GPT model developed in Phase~1, which has already demonstrated strong performance in molecular generation. Here, the training objective is extended to focus on fragment–molecule relationships, where the model is trained on pairs of fragments and their corresponding complete molecules using Causal Language Modeling (CLM). The Phase~2 training corpus is constructed as follows:
\begin{align}\label{eq:phase2_corpus}
\corpus_{\text{Phase 2}}=\curlybracket{[BOS],\tuple{p_1},\underbrace{{f_1}_1,\cdots,{f_1}_{\horizon}}_{\text{fragment $F_1$}},\tuple{p_2},\underbrace{{f_2}_1,\cdots,{f_2}_{\horizon}}_{\text{fragment $F_2$}},\tuple{L},\underbrace{y_1,\cdots,y_{\horizon}}_{\text{target molecule Y}},[EOS]},
\end{align}
where \fix{\tuple{F_1, F_2} \text{: fragments.}} Consequently, the model is able to generate the corresponding complete molecule when provided with a pair of fragments as the input prompt.

\subsection{Surrogate model}\label{app:surrogate_model}
The surrogate model~\citep{vasan23} adopts a simplified BERT-style transformer architecture, widely used in natural language processing. In this framework, tokenized SMILES strings are embedded with positional encodings and passed through a stack of five transformer blocks. Each block consists of a multi-head attention mechanism with 21 heads, followed by dropout, layer normalization with residual connections, and a feedforward subnetwork. The feedforward subnetwork is composed of two dense layers, again followed by dropout and layer normalization with residual connections. After the transformer stack, a final feedforward layer produces the predicted docking score. The validation performance achieved $r^2$ values of 0.73 on the RTCB dataset.

\subsection{\fix{Computing infrastructure \fix{and wall-time comparison}}}\label{app:computing_infrastructure}
We trained our docking surrogate models using 4 nodes 
where each node contains CPUs (64 cores) and 4 A100 GPU nodes
The training time for each model was approximately 3 hours.
We conducted other RL experiments on a cluster that includes CPU nodes (approximately 280 cores) and GPU nodes (approximately 110 Nvidia GPUs, ranging from Titan X to A6000, set up mostly in 4- and 8-GPU configurations). \fix{Based on the computing infrastructure, we obtained the wall-time comparison in \tabref{table:wall-time} as follows.}

 \begin{table*}[ht!]
 \fix{
    \centering
    {\scriptsize
    \scalebox{1}{
    \begin{tabular}{l c c  }
        \toprule
        \textbf{Methods}
        & {\makecell[c]{Total Run Time}}
        \\
        \midrule
        \textbf{\makecell[l]{Pretraining}}
        &  \makecell[r]{36h}
        \\
        \textbf{\makecell[l]{Finetuning}}
        &  \makecell[r]{30h}
        \\
        \bottomrule
    \end{tabular}}}
    \caption{{Wall-time comparison between different methods.} }
        \label{table:wall-time}   
        }
\end{table*}

\subsection{\fix{Hyperparameters and architectures}}\label{app:hyperparameters}
Table \ref{app:tab:hyperparams_pretrain}  provides a list of hyperparameter settings we used for our experiments.

\fix{
For experimentation, we use a set of 1,280 molecules from the RTCB dataset, selecting those with docking scores in the range of $[-14, -6]$, following the range reported in \citep{liu2024erp}. 

When computing the average normalized reward of the original molecules---without incorporating similarity constraints---we assign equal weights of $0.25$ to each of the four objectives: docking, drug-likeness, synthesizability, and solubility.

}

\begin{table*}[h!]
    {
    \centering
    {\scriptsize
    \scalebox{1}{
    \begin{tabular}{c c }
        \toprule
        \textbf{Parameter} &  \textbf{Value} 
        \\
        \midrule
        {\makecell[l]{Pretraining}}
        \\
        \midrule
        {\makecell[l]{\quad Learning rate}} &  \makecell[c]{$5 \times e^{-5}$}
        \\
        \midrule
        {\makecell[l]{\quad Batch size}} &  \makecell[c]{$24$}
        \\
        \midrule
        {\makecell[l]{\quad Optimizer}} &  \makecell[c]{Adam}
        \\
        \midrule
        {\makecell[l]{\quad \# of Epochs for Training First Phase}} &  \makecell[c]{$10$} \\
        \midrule
        {\makecell[l]{\quad \# of Epochs for Training Second Phase}} &  \makecell[c]{$10$} \\
        \midrule
        {\makecell[l]{\quad Model \# of Params}} &  \makecell[c]{$124M$} 
        \\
        \midrule
        {\makecell[l]{Finetuning}}
        \\
        \midrule
        {\makecell[l]{\quad \# of molecules generated $K$}} &  \makecell[c]{$4000$}
        \\
        \midrule
        {\makecell[l]{\quad \# of Epochs $N$}} &  \makecell[c]{$50$}
        \\
        \midrule
        {\makecell[l]{\quad Batch Size}} &  \makecell[c]{$16$}
        \\
        \midrule
        {\makecell[l]{\quad Learning rate}} &  \makecell[c]{$5 \times e^{-5}$}
        \\
        \midrule
        {\makecell[l]{Generation}}
        \\
        \midrule
        {\makecell[l]{\quad TopK}} &  \makecell[c]{$[10,15,20]$}
        \\
        \midrule
        {\makecell[l]{\quad TopP}} &  \makecell[c]{[$0.85$, $0.9$, $0.95$]}
        \\   
        \midrule
        {\makecell[l]{Shared}}
        \\
        \midrule
        {\makecell[l]{\quad Tamimoto Similarity \\ \quad Fingerprint Size}} &  \makecell[c]{1024}
        \\   

        \bottomrule
    \end{tabular}}}
        \caption{{{Hyperparameters for pretraining and generation}}. }
        \label{app:tab:hyperparams_pretrain}
        }
\end{table*}

\end{document}